\documentclass[conference]{IEEEtran}
\IEEEoverridecommandlockouts
\usepackage{cite}
\usepackage{amsmath,amssymb,amsfonts}
\usepackage{algorithmic}
\usepackage{graphicx}
\usepackage{textcomp}
\usepackage{xcolor}
\def\BibTeX{{\rm B\kern-.05em{\sc i\kern-.025em b}\kern-.08em
    T\kern-.1667em\lower.7ex\hbox{E}\kern-.125emX}}
\begin{document}

\title{Chronoamperometry with Room-Temperature Ionic Liquids: Sub-Second Inference Techniques}

\author{\IEEEauthorblockN{Kordel K. France}
\IEEEauthorblockA{\textit{Dept. of Computer Science} \\
\textit{University of Texas at Dallas}\\
kordel.france@utdallas.edu}
}


\pagestyle{plain}

\maketitle

\begin{abstract}
Chronoamperometry (CA) is a fundamental electrochemical technique used for quantifying redox-active species. 
However, in room-temperature ionic liquids (RTILs), the high viscosity and slow mass transport often lead to extended measurement durations. 
This paper presents a novel mathematical regression approach that reduces CA measurement windows to under 1 second, significantly faster than previously reported methods, which typically require 1–4 seconds or longer. 
By applying an inference algorithm to the initial transient current response, this method accurately predicts steady-state electrochemical parameters without requiring additional hardware modifications. 
The approach is validated through comparison with standard chronoamperometric techniques and is demonstrated to maintain reasonable accuracy while dramatically reducing data acquisition time. 
The implications of this technique are explored in analytical chemistry, sensor technology, and battery science, where rapid electrochemical quantification is critical.
Our technique is focused on enabling faster multiplexing of chronoamperometric measurements for rapid olfactory and electrochemical analysis.
\end{abstract}


\section{Introduction}
\vspace{-1mm}

Chronoamperometry (CA) is a potent electrochemical technique where a step change in electrode potential yields a time-dependent current response.
In room-temperature ionic liquids (RTILs) and RTIL-based electrolytes, CA has been widely applied for analyte detection, sensors (including electronic noses for machine olfaction), and battery research. 
RTILs offer wide electrochemical windows and negligible vapor pressure, but their high viscosity and unique double-layer dynamics can slow down diffusion and interfacial equilibration.
These properties often demand longer measurement times to reach diffusion-controlled currents or steady state. 
Researchers have sought methods to shorten the chronoamperometric measurement window without expensive custom hardware. 
Here, we give a brief overview of a technique for implementing shorter CA measurements in RTIL systems and define an algorithmic technique to accelerate or extract information from short transients.
Our method is intended to work without the need for custom hardware and we demonstrate our techniques on the PalmSens Pico MUX16 development board, an off-the-shelf potentiostat.
The intent of our work is to demonstrate how faster detection decisions can be made with electrochemical sensors in biomedical and olfaction applications.


Our contributions can be summarized as follows: 
\begin{enumerate}
\item We demonstrate a technique for inferring the asymptote of the chronoamperometric sequence through short pulses and an inference algorithm.
\item We show how this technique compares to full chronoamperometric sequences.
\item We define the optimal properties for our method and note limits that provide opportunities for future research.
\end{enumerate}

\section{Background}
\vspace{-1mm}
\subsection{Short Measurement Windows in RTIL Chronoamperometry}
Early CA experiments in RTILs often lasted tens of seconds to minutes due to slow faradaic current decay and significant double-layer charging.
For example, in an oxygen-sensing RTIL system, faradaic current did not dominate until $\geq 300$ seconds had elapsed \cite{Lin2018-xz}.
However, several recent studies demonstrate that effective measurements can be made in much shorter time frames.
Wan et al. (2017) introduced a Transient Double-Potential Amperometry (DPA) technique for an RTIL-based gas sensor, enabling reliable oxygen measurements in just 4 seconds \cite{Wan2017-ds}.
By alternately stepping the potential to a reductive sensing value and then to an oxidative value to regenerate the analyte, they significantly reduced the required measurement time while maintaining accuracy. 

Lin et al. (2018) investigated periodic on/off potential steps for an RTIL amperometric $O_2$ sensor \cite{Lin2018-xz}. 
They found that using a sensing pulse as short as 1 second (followed by a longer idle period for recovery) could yield stable sensor signals. 
In their tests, a 1-second polarization period (with a high idle:sensing ratio) produced minimal baseline drift and good repeatability, whereas longer polarization times (e.g. 200 seconds-30 minutes) were more prone to drift.
This demonstrated that even a single second of data, if properly managed and repeated, can be sufficient for analytical purposes in RTIL systems. 
Notably, 1 second was the shortest active measurement period explored in that study.

Very recently, Conceição, et al. (2024) reported a calibration-free quantification method using a single chronoamperometric transient on a microelectrode that “only takes a few seconds” \cite{Conceicao2024-zt}.
Although they do not state an exact minimum time, the implication is that on the order of 2–5 seconds of data were sufficient to determine analyte concentration. 
This is facilitated by the small size of the microelectrode (yielding a fast steady-state current) and advanced data processing (described in the next section). 
Their achievement underscores that, under the right conditions, CA experiments need only last a few seconds to gather meaningful analytical information.
Why not shorter? In theory, microelectrodes in RTILs can reach steady-state currents within milliseconds due to radial diffusion \cite{li2013}.
However, practical limitations (instrument response, double-layer relaxation, and the need to average out noise) often make sub-second CA readings challenging. 
The studies above pushed the boundaries with existing hardware – achieving 1–4 second windows in sensors and “a few seconds” in analytical measurements – all without custom high-speed amplifiers or ultramicrosecond timing. 
We found no prior reports in the RTIL literature explicitly claiming reliable CA quantification in $\leq 1$ second.

\subsection{Software Techniques for Faster Chronoamperometry}

To shorten CA sequences without changing the physical hardware (electrode or potentiostat), researchers have developed clever software and data analysis techniques. 
These approaches extract required information from partial or perturbed transients, effectively speeding up measurements. 
Key techniques are delineated in the following sections:

\subsubsection{Theoretical Curve Fitting and Regression}
One powerful approach is to fit the measured current–time curve to an analytical model, allowing extrapolation beyond the recorded window. 
Xiong, et al. pioneered this by applying the Shoup–Szabo approximation for diffusion-controlled currents \cite{xiong2012}.
The Shoup–Szabo model describes the entire chronoamperometric current decay at a microdisk electrode with <0.5\% error across all times \cite{SHOUP_SZABO_1982}. 
By recording only the initial segment of the CA transient and fitting it to the model, one can deduce key parameters (diffusion coefficient $D$, concentration $c$, or number of electrons $n$) without waiting for long-time asymptotes.
This regression-based strategy essentially compresses a full chronoamperogram’s worth of information into a shorter sampling period, a principle we extend here. 

\subsubsection{Pulsed and Multi-Step Potential Techniques}
Another route to speed up or stabilize short measurements is using specialized potential pulse sequences. 
In double-step or multi-step chronoamperometry, the potential is rapidly toggled to drive reactions and then reverse or relax them. 
Wan et al.’s Double Potential Amperometry (DPA) is a prime example: they applied a brief cathodic step to reduce $O_2$ (measuring the current), immediately followed by an anodic step to oxidize the intermediate back to $O_2$ \cite{Wan2017-ds}.
This pulsed approach accomplishes two things: (1) it shortens the needed cathodic sensing period (they optimized the “oxidation period” and “reduction period” for 4 s total) and (2) it clears out reaction by-products in-between measurements, preventing accumulation that would otherwise necessitate longer waits. 
Lin et al. (2018) used a periodic ON–OFF pulsing of the electrode. By inserting an idle (open-circuit or zero-volt) period after each short sensing pulse, they allowed the RTIL double-layer and reactant distribution to reset.
The net effect is the ability to take rapid successive readings (each on the order of 1–10 seconds) without drift, rather than one prolonged measurement.
Pulsed techniques in general (including classical methods like Differential Pulse Voltammetry or Square-Wave Voltammetry) minimize capacitive currents by sampling after each potential jump \cite{hussain2018}. 
In the context of chronoamperometry, designing sequences that suppress or account for charging currents means one can trust data from earlier time points, shortening the required window.

\subsubsection{Exploiting Diffusion Characteristics}
Fast analysis can also be achieved by focusing on the early-time diffusion behavior. 
In diffusion-limited processes, the CA current follows the Cottrell equation at short times. 
Some studies use this relationship to quickly estimate diffusion rates or concentrations. 
For instance, regression of $i$ versus $t^{-1/2}$ for the initial transient can provide the product $nFAc\sqrt{D}$ (from the slope) almost immediately, which in turn yields concentration if $D$ is known.
In practice, researchers often combine such analysis with microelectrodes – where diffusion reaches steady-state quickly – to get near-instant readings. 
Hussain, et al. (2016) showed that by using microarrays (which enforce radial diffusion) and even “filling” microcavities with deposited platinum to enhance that diffusion, one can obtain higher currents and faster stable responses for gas analytes in RTIL \cite{hussain2016}. 
In summary, techniques that mathematically separate fast and slow processes (capacitive vs. faradaic, or diffusion vs. kinetics) allow experimenters to glean the desired diffusive-charge information from a truncated data set, focusing only on the timeframe where the target process dominates.

In comparison to this recent work, we find that there is further opportunity to prove that RTIL-based CA can be efficiently accomplished within the sub-second time window.
We believe our work to be an early first demonstration of low sub-second chronoamperometry in RTILs.
Our utilization of inference techniques to extract electrochemical parameters from partial transients require no hardware modifications, making it broadly applicable to all existing chronoamperometric sensors.

\section{Methodology}
\vspace{-1mm}

\begin{figure}
  \centering
  \includegraphics[width=85mm]{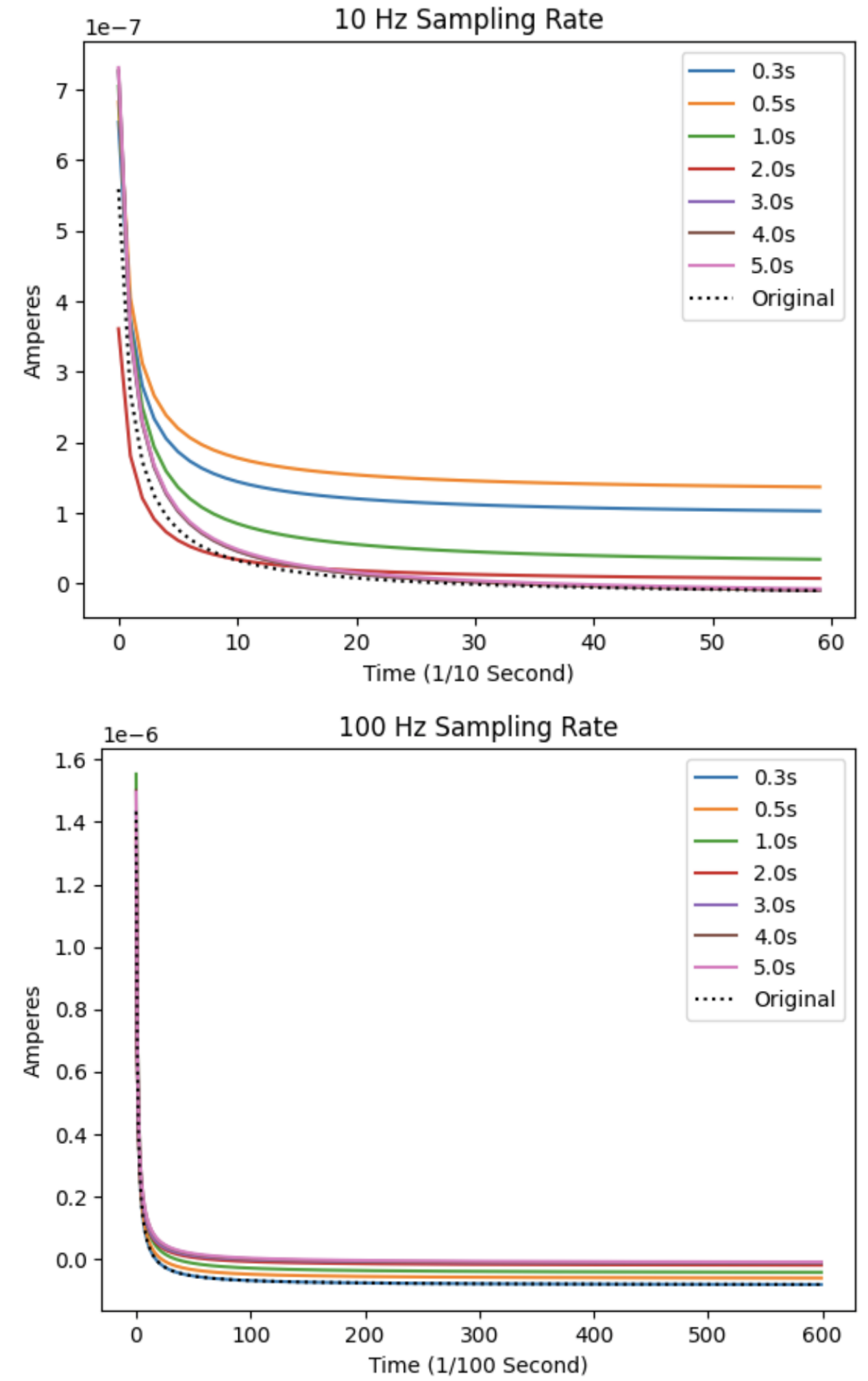}
  \caption{Examples of inference plots showing predictions of the first 6 seconds over different values of \textit{n}. Both 10 Hz (top) and 100 Hz (bottom) chronoamperometry resolutions were analyzed.}
  \label{fig:chronoampRegressions}
\end{figure}

Chronoamperometry employs rapid potential steps to induce electrochemical reactions, generating current-time responses. 
Fast sampling is crucial for capturing transient signals and resolving rapid kinetics. 
The current generates at early time domain is depicted as Transient Diffusion Current (TDC). 
Transient diffusion current arises when charge carriers redistribute in response to sudden changes in concentration gradients or electric fields. 
Many metabolites, having volatile property diffuse through the electrode double layer, creating the gradient, which generates the TDC, occurs between 5s-7s \cite{Banga2022}.
We focus demonstrating our inference technique over this time measurement window as a result to focus the feasibility study here.
Chronoamperometry is governed by the Cottrell equation: 

\begin{equation}
    I = \frac{n_eFAc_k \sqrt{D_k}}{\sqrt{\pi t}}
    \label{eq:cottrell}
\end{equation}

\noindent where $I$ denotes the electrical current, measured in amperes; $n_e$ is the number of electrons needed to oxidize one molecule of analyte $k$; $F$ is the Faraday constant of 96,485 Coulombs per mol; $A$ denotes the planar area of the electrode in square centimeters; $c_k$ defines the initial concentration of the target analyte $k$ in mols per cubic centimeter; $D_k$ defines the diffusion coefficient for analyte $k$ in square centimeters per second; and $t$ is simply the time the chronoamperometric sequence is running in seconds. 

We focus on one compound, toluene, in this manuscript for brevity and experimental control.
Toluene has a reduction potential of 1.05V with the 1-Ethyl-3-methylimidazolium tetrafluoroborate (EMIM-BF4) RTIL.
For the purposes of our sensor which contains an electrode surface area of 2.25 square centimeters, according to Fick's law, Equation \ref{eq:cottrell} reduces to:

\begin{equation}
     I = \frac{n_e * 96,485 * 2.25 * c_k * \sqrt{D_k}} {\sqrt{\pi * 1.5}} \approx 1e^{5}n_e c_k \sqrt{D_k}
     \label{eq:cottrellreduced}
\end{equation}

The more time the analyte has to diffuse across the electrode, the more accurate the current response from the sensor will be. 
We set $t=6.0$ and run a normal chronoamperometry sequence to establish the ground truth 6-second baseline.
We note this sequence as the \textit{baseline sequence}.
To evaluate the inference algorithm, we follow the same procedure but terminate the sequence after only $k$ seconds.
Each of these sequences are noted as the \textit{inference sequences}.
We then use an inference algorithm over this data to predict the 6.0-second diffusion value for each sampled measurement.
We desire to find an algorithm that harmonizes (1) simplicity, (2) accuracy, and (3) a low computational budget. 
Upon previous evaluation of several different regression models, we find that inverse regression intersects these three qualities well.
The equation for inverse regression is as follows:

\begin{equation}
    k_i = u + \frac{v}{b_i}
    \label{eq:inverse_regression}
\end{equation}

\noindent Where $b_i$ represents the $i^{th}$ value for the baseline measurement, $v$ represents the gradient of the measurement, $u$ denotes the bias, and $k_i$ is the computed (or inferred) value from each of these parameters.


This value for $t$ governs how long the chronoamperometric sequence is run, but not how many times we sample from the electrode running the sequence.
For this sampling frequency, we select 10 Hz and 100 Hz to evaluate.
Electrochemical signals are highly influenced by aerodynamic factors such as temperature, pressure, and humidity.
To weight the signal according to these factors, we "correct" them according to the following correction equations:

\begin{equation}
    x_{temp}=(0.148*x_{raw})+25.478e^{2.694e^{-9}} k_{rh}-25.478 		
    \label{eq:correction_temp}
\end{equation}
\vspace{-2mm}
\begin{equation}
    x_{rh}=(0.148*x_{temp})-0.999e^{-2.233e^{-9}} k_{temp}+0.999
    \label{eq:correction_rh}
\end{equation}

\noindent
where  $k_{rh}$ denotes the relative humidity of the air prior to correction and $k_{temp}$ denotes the temperature of the air prior to correction. 
The raw chronoamperometric signal $x_{raw}$ is corrected for temperature by Equation \eqref{eq:correction_temp} to give $x_{temp}$.
This value is then corrected for relative humidity by Equation \eqref{eq:correction_rh} to give $x_{rh}$, which is then used as the final reportable value in our results.
These equations are experimentally derived using regression techniques over a battery of tests performed in the laboratory under stepped pressure, temperature, and humidity levels.
Typically, one must also adjust these values for pressure, but we found that the pressure correction equation provided little impact to our sensing environment, so we relieve it from our methodology in order to reduce computational complexity.
\vspace{-1mm}

\begin{table*}
    \caption{Variability Statistics of Inferred Diffusion Currents}
    \begin{center}
    \begin{tabular}{|c|c|c|c|c|c|c|}
    \hline
    \textbf{Sampling Rate (Hz)}& \textbf{Inter-sequence $\mu$ (nA)} & \textbf{Inter-sequence $\sigma$(nA)}& \textbf{Delta $\mu$ (nA)} & \textbf{Delta $\sigma$ (nA)}& \textbf{Average $R^2$}\\
    \hline
    10& 68.596& 105.869& 46.4642& 55.415& 0.844\\
    100& 43.406& 2790.235& 1013.969& 26.138& 0.964\\
    \hline
    \end{tabular}
    \label{tab:resultsTable}
    \end{center}
\end{table*}
\vspace{-3mm}

\section{Experimental Setup \& Results}
\vspace{-1mm}
Our inference technique was evaluated over two different sampling rates (10 Hz and 100 Hz) and seven different time sequences: 0.3, 0.5, 1.0, 2.0, 3.0, 4.0, and 5.0 seconds.
The 6.0-second sequence is the true measurement of the full chronoamperometric sequence while the other seven sequences are inferred using our inference algorithm.
We dropcast 1.5 microliters of the RTIL onto the surface of the electrode and spincoat at 1200 RPMs to ensure consistent distribution of the RTIL across the electrode surface.
The RTIL was purchased from Sigma-Aldrich and the electrode was purchased from Texas Instruments.
Our sample of RTIL is at 120 part-per-billion (ppb) concentration. 
For each experiment, we expose the vial to the ambient air 500 mm from the electrode before we run the chronoamperometric sequence.
After each sequence, we vent the surrounding air to remove the stimulus (toluene) as reasonable as possible and then wait 5 minutes before we perform the next sequence, borrowing the "relaxation period" from \cite{Lin2018-xz}.
We take five measurements at each time sequence and average them.

The results of the final inferred chronoamperometric curves are displayed in Figure \ref{fig:chronoampRegressions}.
Statistics regarding the variability over the inferred curves are shown in Table \ref{tab:resultsTable}
\footnote{A more comprehensive list of results are available upon request.}.
For each sampling rate, we compute the mean and standard deviation of the variability between all inferred sequences and the baseline sequence (Inter-sequence $\mu$ \& $\sigma$). 
We also compute the average correlation between each inferred sequence to the baseline (Average $R^2$) as well as the mean and standard deviation of the delta between the final inferred and baseline transient diffusion currents (Delta $\mu$ \& $\sigma$).

Over both sampling rates, we find that the inference sequences at the lower 10 Hz sampling rate has lower variability but also lower overall correlation to the baseline sequence.
The higher 100 Hz sampling rate has higher variability but a higher overall average correlation to the baseline sequence over all inference sequences.
For the each sampling rate, we take the inference sequence whose line provides the highest overall correlation and smallest difference between the inferred transient diffusion current and the true baseline diffusion current.
For the 10 Hz sampling rate, we find that the 5.0-second inference window achieves the highest correlation to the baseline sequence with the lowest error.
The inference line of best fit for 10 Hz sampling rate is
\begin{equation}
    k_i=-2.0114e^{-8} + 7.5093e^{-7}/b_i
        \label{eq:tenHzFinal}
\end{equation}
\vspace{-1mm}

\noindent with correlation 88.55\%.
However, this does not accomplish our goal of sub-second chronoamperometry.
Over the 100 Hz sampling rate, we find that the 0.3-second measurement window achieves the highest correlation to the baseline sequence with the lowest error.
This is much more aligned with our goals of fast chronoamperometry inferred via sub-second pulses.
 The inference line of best fit for 100 Hz sampling rate is
\begin{equation}
    k_i=-8.4844e^{-8} + 1.5182^{e-6}/b_i
    \label{eq:hundredHzFinal}
\end{equation}
\vspace{-1mm}

\noindent with correlation 97.68\%.



In summary, our evaluation of our inference model over several measurement windows yields compelling results only when sampling at 100 Hz.
In other words, we find we can achieve predictions of the 6.0-second transient diffusion current from a 0.3-second pulse, but only when sampling at 100 Hz.
Equation \ref{eq:hundredHzFinal} gives the final inference model that can be used to infer each additional measurement value for the 6.0-0.3=5.7 seconds from the initial 0.3 second pulse.
\section{Conclusion}
\vspace{-1mm}
The study above demonstrates early work in inferring full-sequence transient diffusion currents via sub-second pulses in chronoamperometry.
While our methodology shows promise over toluene in laboratory conditions, there are still many uncertainties left to be resolved to ensure the method extrapolates to other compounds and scales to more scenarios.
Future work will investigate compounds from other chemical groups (e.g. aromatics, aliphatics, etc.) over a wider array of environmental conditions.
We strive to make our work accessible and replicable by conducting our experiments on off-the-shelf hardware and low-computation algorithms and hope this encourages more work in rapid electrochemical techniques closer to those of sub-second measuring intervals.
\vspace{-1mm}

\vspace{-1mm}
\centering
\bibliographystyle{IEEEtran}
\bibliography{sample-base}

\end{document}